\pdfoutput=1
\documentclass[11pt]{article}

\usepackage[]{emnlp2021}

\usepackage{times}
\usepackage{latexsym}

\usepackage[T1]{fontenc}
\usepackage[utf8]{inputenc}

\usepackage{microtype}

\usepackage{graphicx}
\usepackage{booktabs}
\usepackage{multirow}
\usepackage{rotating}
\usepackage{amsmath}
\usepackage{amssymb}
\usepackage{bm}
\usepackage{url}
\usepackage{CJKutf8}
\usepackage{floatrow}
\usepackage{xcolor}
\usepackage{tabu}
\newfloatcommand{capbtabbox}{table}[][\FBwidth]

\title{EVA: An Open-Domain Chinese Dialogue System with Large-Scale Generative Pre-Training}

\author{Hao Zhou\thanks{\quad Equal contribution}, Pei Ke$^*$, Zheng Zhang$^*$, Yuxian Gu, Yinhe Zheng, Chujie Zheng, \\ \textbf{Yida Wang, Chen Henry Wu, Hao Sun, Xiaocong Yang, Bosi Wen,}  \\ \textbf{Xiaoyan Zhu, Minlie Huang\thanks{\quad Corresponding author: Minlie Huang. Email address: aihuang@tsinghua.edu.cn}, Jie Tang} \\
Department of Computer Science and Technology,
Tsinghua University \& BAAI \\
}

\begin{document}
\maketitle
\begin{abstract}

Although pre-trained language models have remarkably enhanced the generation ability of dialogue systems, open-domain Chinese dialogue systems are still limited by the dialogue data and the model size compared with English ones. In this paper, we propose EVA, a Chinese dialogue system that contains the largest Chinese pre-trained dialogue model with 2.8B parameters. To build this model, we collect the largest Chinese dialogue dataset named WDC-Dialogue from various public social media. This dataset contains 1.4B context-response pairs and is used as the pre-training corpus of EVA. Extensive experiments on automatic and human evaluation show that EVA outperforms other Chinese pre-trained dialogue models especially in the multi-turn interaction of human-bot conversations\footnote{The codes, data, and model parameters will be available at \url{https://github.com/thu-coai/EVA}.}.

\end{abstract}

\section{Introduction}

%% data group
In recent years, numerous focus has been investigated to build open-domain dialogue systems, which require generating responses based on users' input posts in open domains. Early works on open-domain dialogue systems mainly depend on RNN-based sequence-to-sequence (Seq2Seq) models \cite{vinyals2015seq2seq,shang2015nrm}. With the development of pre-trained language models such as GPT \cite{radford2018gpt}, BART \cite{lewis2020bart} and T5 \cite{raffel2020t5}, 
latest works in this area resort to building open-domain dialogue systems based on large-scale generative pre-training models, which include DialoGPT \cite{zhang2020dialogpt}, Meena \cite{adiwardana2020meena} and Blender \cite{roller2021blender}. Equipped with large amounts of dialogue data collected from social media, these models can generate human-like responses and improve the engagingness of human-bot conversations.

However, most of the dialogue models based on large-scale pre-training are built in English. We argue that existing works on open-domain Chinese dialogue systems are limited in model and data sizes.
For example, CDial-GPT \cite{wang2020lccc} (with 104M parameters) is pre-trained on 12M Chinese dialogues from Weibo\footnote{\url{https://weibo.com/}}. PLATO-2 \cite{bao2020plato2} (with 336M parameters) is pre-trained on 1.2B Chinese dialogues from social media. The scale of the publicly available dialogue data hinders us from building Chinese pre-trained dialogue models that can generate high-quality responses on open-domain topics.
% Further, the scale of model parameters degrades the performance on dialogue understanding and generation.

\begin{table*}[tbp]
\centering
    \scalebox{0.87}{
\begin{tabular}{@{}lcccccc@{}}
\toprule
{Dataset}             & {\#Sess}. & {\#Utter.} & {\#Token} & \begin{tabular}[c]{@{}c@{}}{Avg. \#utter.}\\ {per sess.}\end{tabular} & \begin{tabular}[c]{@{}c@{}}{Avg. \#token}\\ {per utter.}\end{tabular} & {Storage size} \\ \midrule
% \textbf{Dataset}             & \textbf{\# Sess}. & \textbf{\# Utter.} & \textbf{\# Token} & \begin{tabular}[c]{@{}c@{}}\textbf{Avg. \# utter.}\\ \textbf{per sess.}\end{tabular} & \begin{tabular}[c]{@{}c@{}}\textbf{Avg. \# token}\\ \textbf{per utter.}\end{tabular} & \textbf{Storage size} \\ \midrule
LCCC-base \cite{wang2020lccc}   & 6.8M     & 20.0M     & 232.3M   & 2.9                            & 11.6       & 911MB        \\
LCCC-large \cite{wang2020lccc}  & 12.0M    & 32.9M     & 380.1M   & 2.7                            & 11.6          & 1.5GB        \\
PLATO-2 \cite{bao2020plato2}    & 1.2B     & -         & -        & -                                         & -        & -            \\
STC \cite{shang2015nrm}       & 4.4M     & 8.9M      & 158.1M   & 2                                         & 25.2    & 642MB        \\
Douban Conversation \cite{wu-etal-2017-sequential} & 1.0M     & 7.1M      & 131.7M   & \textbf{6.7}   & 18.6        & 535MB        \\
PersonalDialog \cite{zheng2019personalized}      & 20.8M    & 56.2M     & 525.9M   & 2.7          & 9.4     & 2.1GB        \\
PchatbotW    \cite{qian2021pchatbot}       & 139.0M   & 278.9M    & 8.5B     & 2                  & \textbf{30.5}       & 50GB         \\
PchatbotL   \cite{qian2021pchatbot}        & 59.4M    & 118.9M    & 3.0B     & 2                   & 25.5                    & 19GB         \\
\textbf{WDC-Dialogue (Ours)} & \textbf{1.4B}     & \textbf{3.0B}      & \textbf{78.3B}    & 2.1            & 26.2    & \textbf{181GB}        \\ \bottomrule
\end{tabular}
}
\caption{Statistics of WDC-Dialogue and existing Chinese dialogue datasets. - means that the value is not reported in the original papers.}
\label{tab:data-stat}
\end{table*}

In this paper, we build an open-domain Chinese dialogue system called \textit{EVA}, which contains the largest Chinese dialogue model with 2.8B parameters and is pre-trained on WDC-Dialogue, including 1.4B Chinese dialogue data from different domains. First, we construct the WDC-Dialogue dataset by collecting the repost, comment, and Q\&A data from various social media platforms and refactor them into dialogue sessions. Strict filtering rules are also devised to ensure the quality of the WDC-Dialogue dataset. Second, we train a large-scale Transformer-based encoder-decoder model on the Chinese dialogue data. To verify the effectiveness of our model, we conduct extensive automatic evaluation and human evaluation. In the automatic evaluation, we test our model on four datasets to show the generation ability when dealing with different categories of contexts. Moreover, observational and interactive human evaluations are also adopted to evaluate our model in real human-bot conversation scenarios. Finally, we provide an interactive demonstration system for users to converse with EVA.

Our contributions are mainly as follows:

\begin{itemize}
    \item We collect the largest Chinese dialogue dataset called WDC-Dialogue from different domains, which contains 1.4B context-response pairs. The data quality is controlled by strict rules.
    \item We build an open-domain dialogue system called EVA 
    %with 2.8B model parameters
    , which contains the largest Chinese pre-trained dialogue model with 2.8B parameters.
    %which is the largest Chinese dialogue system. 
    Extensive experiments on automatic and human evaluation show the effectiveness of our model.
    \item We release an interactive demonstration system for users to converse with EVA on open-domain topics.
\end{itemize}

\section{Data}
We construct a dataset named \textit{WDC-Dialogue} from Chinese social media to train EVA.
Specifically, conversations from various sources are gathered and a rigorous data cleaning pipeline is designed to enforce the quality of WDC-Dialogue.
This section details the data collection and cleaning process used in our study.

\begin{table*}[]
    \centering
    \small
    \begin{tabular}{ccccccc}
    \toprule
                                   & $n_{param}$ & $L$ & $n_{head}$ & $d_{model}$ & $d_{ff}$ & Model Type   \\
    \midrule
        % Plato                      & 110M        & 12  &    12      & 768         & 3,072    & UniLM        \\
        CDial-GPT                  &  104M       & 12  &    12      & 768         & 3,072    & Decoder         \\
        PLATO-2                    &  336M       & 24  &    16      & 1,024       & 4,096    & UniLM               \\
        EVA                        &  2.8B       & 24  &    32      & 2,048       & 5,120    & Encoder-Decoder     \\
    \bottomrule
    \end{tabular}
    \caption{Comparison between EVA and other large-scale Chinese pre-trained dialogue models.}
    \label{tab:models}
\end{table*}

%% data group
\subsection{Data Collection}
Dialogues in the WDC-Dialogue dataset originate from the textual interaction among different users on the Internet.
Generally, these interactions can be classified into three categories:
1) The interactions exhibited through the repost behaviour on social media;
2) The interactions established through the comment / reply action on various online forums;
3) The interactions about online question and answer (Q\&A) exchanges.
Each round of these textual interactions yields a dialogue session.
We design specific parsing rules to extract dialogues from these three kinds of interactions.
%In our study, specified parsing rules are designed to extract these three kinds of conversations:

\paragraph{Repost}
is a common feature provided by most social media platforms, which allows users to broadcast the posts created by others and add their own comments to these original posts (such as the Quote Tweet feature on Twitter).
Each repost can be further broadcast by other users, thereby forming a chain of the user reply. This chain can be refactored in a dialogue session.
%Thus chain of comments can be regarded as a session of dialogue.

In practice, we observe that such interaction pattern yields a \textit{reply tree},
in which the root node is the original post, and the other nodes consist of the comments added in the broadcasting process.
Each node may have multiple child nodes, which denote the comments left in the reposting process when broadcasting this node.
Once the reply tree is constructed, each path from the root to the leaf can be regarded as a dialogue session.

In this study, we target several Chinese social platforms.
Specifically, the raw data of reposts are first collected and further parsed to construct the reply trees.
Dialogues are obtained using a Depth-First-Search algorithm to traverse all the paths from each root node to their leaf nodes.
This process helps to collect dialogues containing multiple turns of interaction.

\paragraph{Comment}
is another common feature that facilitates textual interactions among different users who surf the Internet.
It allows users to share their opinion by leaving textual comments, which can be further replied to by others.
Such an interaction pattern can be regarded as a form of conversation among users.

In this study, we target various Chinese forums.
The raw data of posts and their following comments are collected.
Ideally, these raw data can also be parsed to form a reply tree because each comment may have multiple replies.
However, compared with the repost data, the collection of comment data is less flexible because some of the HTML pages from the front-end interface of forums do not provide the detailed reply information of each comment. As a consequence, the depth of reply trees is limited and the dialogues obtained based on comment data have shorter turns comparing to dialogues originating from repost data.

%However, unfortunately, the HTML pages obtained from the front-end interface of each forum do not provide the detailed reply relation of each comments, and thus the ``reply tree'' can not be constructed. We observe that the best we can do is to construct dialogues that contain at most three utterances.

\paragraph{Q\&A}
is a special kind of interaction among users on the Internet.
%The forums such as \textit{Quora} or \textit{Zhihu} based data is a popular form in current social network sites.
In online Q\&A platforms such as \textit{Quora}\footnote{\url{https://www.quora.com/}} or \textit{Zhihu}\footnote{\url{https://www.zhihu.com/}}, users post their questions related to various topics attached with a detailed description. Other people tend to provide answers with lots of backgrounds, opinions, experiences, and knowledge. We regard a post and each of its corresponding answers as a single-turn conversation.

\subsection{Data Quality Control}

Textual data from online social media carry various noises such as advertisements, hate speech, profanity and informal internet slang. Some of the contents even carry sensitive information such as user privacy. Models trained on these data can easily bias to these noisy contents. To improve the quality of the WDC-Dialogue dataset, we design a rigorous process to clean the dialogues.

We follow a similar process used by \citet{wang2020lccc} to filter out noisy contents with a series of rules: (1) delete the platform-related tags in the dialogues, such as "Reply to @***" and "Repost//@***"; (2) remove URL strings from the text; (3) split conversations with more than 30 turns into multiple conversations less than 30 turns \citet{shang2015nrm}; (4) only keep one copy of the phrases or words that repeat more than 6 times in one sentence; (5) remove dialogues that contain responses that are too long or too short; (6) remove dialogues if the response is identified as an advertisement by the method introduced in \citet{wang-etal-2013-dataset}; (7) remove dialogues if 90\% of tri-grams in the response are high-frequency tri-grams \cite{zhang2020dialogpt}; (8) remove dialogues if the response has some specific forms of generic responses; (9) remove dialogues in which the response is the same as the post.

We also manually construct a word list containing the following noise: (1) dirty words, sensitive words, and dialect; (2) special topic words such as the name of some rare virus or compound; (3) name, appellation and unknown abbreviation; (4) special symbols and emojis; (5) platform signs such as the words which are related to ads, pictures, and videos. A dialogue will be removed from our dataset if it contains words in this word list.

\subsection{Data Statistics}

Table \ref{tab:data-stat} shows a statistics of the filtered WDC-Dialogue dataset and other Chinese dialogue datasets. To the best of our knowledge, WDC-Dialogue is the largest Chinese dialogue dataset with 1.4B context-response pairs and the largest number of utterances, tokens and storage size.
%It consists of xxxx single-turn and xxxx multi-turn sessions. 
% As shown in Table \ref{tab:data-stat}, WDC-Dialogue contains the .

\section{Method}

%%% model group

\subsection{Model}

EVA is a Transformer-based dialogue model with a bi-directional encoder and a uni-directional decoder \cite{VaswaniSPUJGKP17}. We present the EVA's model details and a comparison with previous large-scale Chinese pre-trained dialogue models in Table~\ref{tab:models}. EVA is nearly 8 times the size of the previous largest Chinese dialogue model, PLATO-2. 

\subsection{Tokenization}

As Chinese words, containing some specific meanings, are usually composed of several characters, traditional character-level vocabulary loses the important semantics of Chinese words or phrases. Thus, we construct a sub-word vocabulary, containing both Chinese characters and Chinese words, based on the word segmented corpus using unigram language model~\cite{DBLP:conf/emnlp/KudoR18}. The sub-word vocabulary contains 30,000 tokens.

\subsection{Pre-Training Details}

We use the sequence-to-sequence language modeling \cite{SutskeverVL14} task to train our model. Specifically, for a dialogue session with $n$ utterances, we train the model to generate the $n^{th}$ utterance from the decoder conditioned on the previous $n-1$ utterances, which are fed to the encoder. The model is trained with the teacher-forcing paradigm.

To reduce the GPU memory consumption, we adopt mixed-precision training~\cite{micikevicius2018mixed} and ZeRO~(stage-1)~\cite{rajbhandari2020zero} to partition the parameters of the optimizer to multiple data parallelism process. 

\begin{figure}[t]
	\centering
	\includegraphics[width=0.9\linewidth]{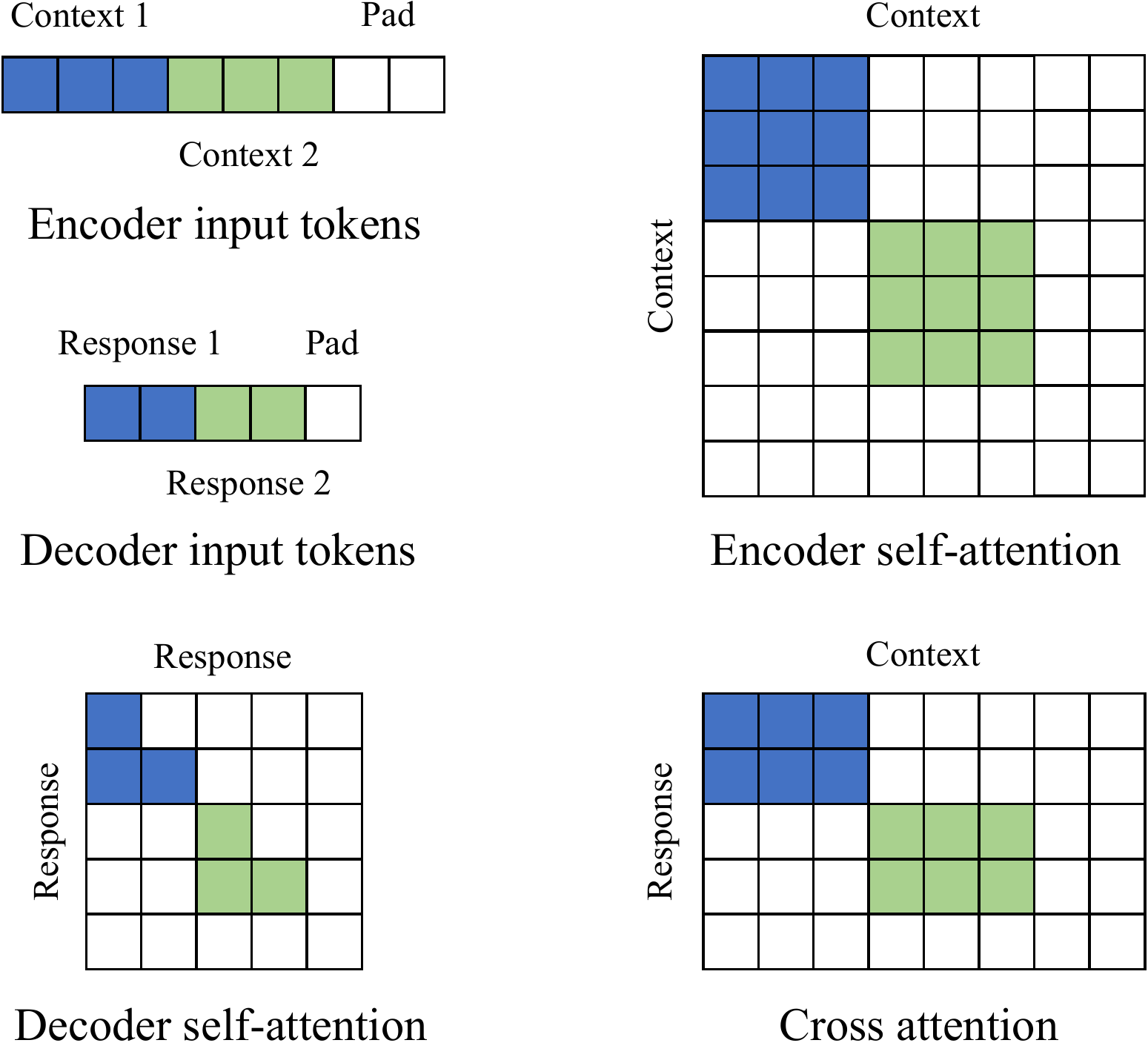}
	\caption{\label{fig:data-sampling} Our data sampling strategy for efficient pre-training. In this example, two context-response pairs are concatenated and padded as a data sample. With the attention masks, the two pairs cannot attend to each other. The relative position embeddings are compatible with our sampling strategy. 
	}
\end{figure}

We set the maximum encoder length and maximum decoder length as 128 to ensure that most utterances are not truncated during training. However, short utterances are heavily padded if we view each context-response pair as a data sample; such heavy padding is a bottleneck in pre-training efficiency. To address the challenge, we propose a data sampling strategy that allows a data sample to contain multiple context-response pairs, as illustrated in Figure~\ref{fig:data-sampling}. Specifically, we concatenate multiple context-response pairs as a data sample and distinguish different pairs with attention masks for the encoder self-attention, decoder self-attention, and cross attention. Note that EVA adopts relative position embeddings \cite{raffel2020t5}, which is compatible with our data sampling strategy.

\begin{table} [!htp]
\centering
\footnotesize
\setlength{\tabcolsep}{1.0mm}{
\begin{tabu}{lcccc}
\toprule
%Test Set & Single & Multi & Long & QA \\
Test Set & \#Inst. & \#Uttr. & \#Token (Cont.) & \#Token (Resp.) \\
\midrule
\textbf{Single} & 10,000 & 2.0 & 16.8 & 11.5 \\
\textbf{Multi} & 10,000 & 3.1 & 24.3 & 9.6 \\
\textbf{Long} & 10,000 & 2.0 & 8.9 & 15.9 \\
\textbf{QA} & 10,000 & 2.0 & 22.1 & 12.3 \\
\bottomrule
\end{tabu}}
\caption{Statistics of four test sets, including the amount of instances, the average number of utterances in dialogue sessions, and the average number of tokens in contexts / responses.}
\label{tab:testdatastat}
\end{table}

\section{Experiment}

\subsection{Dataset}

We collect four datasets which have no overlap with our pre-training corpus to test pre-trained dialogue models in a zero-shot setting. These test sets indicate the following dialogue scenarios: 1) \textbf{Single}: This test set contains the dialogue with only one utterance as the context. 2) \textbf{Multi}: This test set includes the dialogue with multiple utterances as the context. 3) \textbf{Long}: This test set contains the dialogues where the length of responses is longer than that of contexts. 4) \textbf{QA}: This test set includes the dialogues where the last utterance of contexts is a question. The statistics of these four test sets are shown in Table \ref{tab:testdatastat}.

\subsection{Baseline}

We adopt several Chinese pre-trained models as our baselines:

%\begin{itemize}

\noindent\textbf{CDial-GPT}: This Chinese pre-trained dialogue model with 104M parameters is pre-trained on LCCC, which contains 12M dialogue sessions \cite{wang2020lccc}.
    
\noindent \textbf{CPM}: This model is a general Chinese pre-trained model with 2.6B parameters, which is pre-trained on 100GB Chinese data including encyclopedia, news, novels, and Q\&A \cite{zhang2020cpm}. Since CPM cannot be directly applied to generating responses for dialogue contexts, we follow the original paper to condition the language model on a prompt of several example context-response pairs.

%\end{itemize}

% \noindent \textbf{CDial-GPT}: This Chinese pre-trained dialogue model with 104M parameters is pre-trained on LCCC, which contains 12M dialogue sessions \cite{wang2020lccc}.

% \noindent \textbf{CPM}: This model is a general Chinese pre-trained model with 2.6B parameters, which is pre-trained on 100GB Chinese data including encyclopedia, news, novels, and Q\&A \cite{zhang2020cpm}. Since CPM cannot be directly applied to generating responses for dialogue contexts, we follow the original paper to condition the language model on a prompt of several example context-response pairs.

Note that we do not choose PLATO-2 \cite{bao2020plato2} as our baseline because the authors have not released the Chinese pre-trained dialogue model.

\subsection{Automatic Evaluation}

%In automatic evaluation, the models generate responses given the contexts in the test sets. Specifically, we evaluate models' performance on 4 different dialogue scenarios (each corresponding to a test subset):

%\noindent \textit{\textbf{Single}} \quad The context contains only one utterance.

%\noindent \textit{\textbf{Multi}} \quad The context contains multiple utterances.

%\noindent \textit{\textbf{Long}} \quad The length of the response is longer than that of the context.

%\noindent \textit{\textbf{QA}} \quad The last utterance in the context is asking questions.

We adopt unigram F1 \cite{dinan2019wizard}, ROUGE-L (R-L) \cite{lin-2004-rouge}, BLEU \cite{papineni-etal-2002-bleu} and Distinct n-grams (Dist-n) \cite{li-etal-2016-diversity} as automatic metrics.
The former three metrics evaluate the relevance between the generated responses and the references, while the last one measures the diversity of generated responses.

The results on the four test sets and the overall results are provided in Table \ref{tab:auto}.
We can see that EVA outperforms both competitors on the relevance metrics, which shows that our model can generate high-quality responses that have more overlap with human references.
EVA also surpasses CDial-GPT in terms of diversity, while performing worse than CPM. We conjecture that the high diversity of CPM may result from more diverse pre-training corpora (rather than the mere dialogue corpus adopted by EVA). We also observe that EVA achieves relatively stable performance on four test sets, indicating that EVA can deal with different kinds of contexts.

%As for the performance of EVA on four test sets, we can observe that EVA achieves the best performance on the relevance metrics including ROUGE-L and BLEU on the Single test set, which indicates that single-turn dialogues are easier for EVA to handle.

%Specifically, EVA achieves better performance on the relevance metrics but worse performance in terms of diversity in the \textbf{Single} test set than the \textbf{Multi} test set, as multi-turn conversations provide opportunities for conversational models to generate diverse responses. Compared to the \textbf{Long} test set, EVA performs better in the  \textbf{QA} test set in terms of relevance, demonstrating that ...

\begin{table}[t]
  \centering
  \scalebox{0.76}{
    \begin{tabular}{llccccc}
    \toprule
    Test Set & Model & F1    & R-L   & BLEU-4  & \multicolumn{2}{c}{Dist-2/3} \\
    \midrule
    \multirow{3}[2]{*}{\textbf{Single}} & CDial-GPT & 6.5   & 5.9   & 0.68  & 18.0  & 36.4  \\
          & CPM   & 8.2   & 7.5   & 1.26  & \textbf{29.0} & \textbf{58.4} \\
\cmidrule{2-7}          & EVA   & \textbf{10.5} & \textbf{9.8} & \textbf{1.88} & 23.6  & 49.5  \\
    \midrule
    \multirow{3}[2]{*}{\textbf{Multi}} & CDial-GPT & 6.4   & 5.9   & 0.55  & 17.0  & 34.2  \\
          & CPM   & 7.7   & 7.2   & 1.01  & \textbf{30.5} & \textbf{60.2} \\
\cmidrule{2-7}          & EVA   & \textbf{10.1} & \textbf{9.5} & \textbf{1.22} & 22.8  & 49.3  \\
    \midrule
    \multirow{3}[2]{*}{\textbf{Long}} & CDial-GPT & 6.5   & 5.6   & 0.28  & 17.4  & 33.4  \\
          & CPM   & 8.7   & 7.6   & 0.78  & \textbf{29.1} & \textbf{57.5} \\
\cmidrule{2-7}          & EVA   & \textbf{10.8} & \textbf{9.6} & \textbf{1.19} & 22.5  & 47.0  \\
    \midrule
    \multirow{3}[2]{*}{\textbf{QA}} & CDial-GPT & 7.2   & 6.5   & 0.59  & 18.4  & 38.9  \\
          & CPM   & 8.2   & 7.3   & 1.12  & \textbf{26.4} & \textbf{55.2} \\
\cmidrule{2-7}          & EVA   & \textbf{10.0} & \textbf{9.1} & \textbf{1.38} & 22.0  & 48.0  \\
    \midrule
    \multirow{3}[2]{*}{\textbf{Overall}} & CDial-GPT & 6.6   & 6.0   & 0.53  & 11.3  & 27.3  \\
          & CPM   & 8.2   & 7.4   & 1.07  & \textbf{18.7} & \textbf{45.3} \\
\cmidrule{2-7}          & EVA   & \textbf{10.4} & \textbf{9.5} & \textbf{1.53} & 14.0  & 35.9  \\
    \bottomrule
    \end{tabular}%
  }
  \caption{Results of automatic evaluation.}
  \label{tab:auto}%
\end{table}%

\subsection{Observational Human Evaluation}

Since automatic metrics cannot completely reflect the generation quality \cite{liu2016hownot}, we further conduct observational human evaluation. We follow the existing work \cite{adiwardana2020meena} to adopt sensibleness, specificity, and the average of sensibleness and specificity (SSA) as our evaluation metrics.
Specifically, sensibleness $\in \{0, 1\}$ measures whether the response is fluent and readable and is coherent to the context.
Specificity $\in \{0, 1\}$ measures whether the responses is specific and informative. Note that the specificity score will be set to 0 if the sensibleness score is 0.
We randomly sampled 200 dialogues from the test set, where each dialogue is judged by 3 annotators.

\begin{figure}[t]
  \centering
  \includegraphics[width=\linewidth]{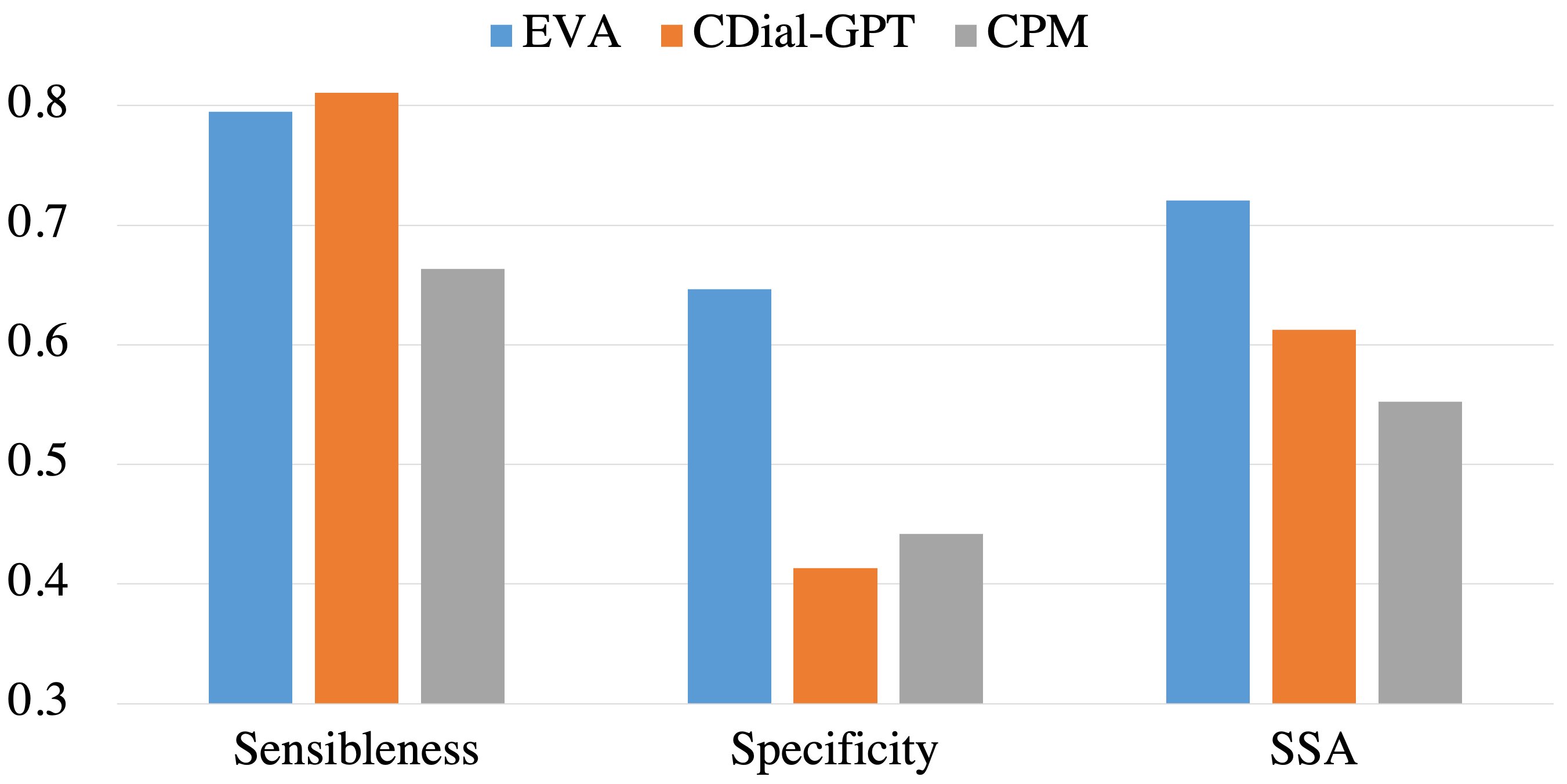}
  \caption{
  Results of observational human evaluation.
  }
  \label{fig:obs}%
\end{figure}

The results are shown in Figure \ref{fig:obs}.
We can see that EVA achieves remarkably higher specificity and SSA scores than baselines, and obtains a comparable sensibleness score to CDial-GPT. It demonstrates that EVA can generate more specific informative responses while maintaining good fluency and contextual coherence. By contrast, CDial-GPT tends to generate safe and generic responses, which are fluent in most cases while containing unspecific meanings. Thus, it obtains the lowest specificity among the three models.

\subsection{Interactive Human Evaluation}

We also follow the existing work \cite{adiwardana2020meena} to conduct interactive human evaluation to simulate the real scenarios of human-bot conversations. We adopt the same metrics as the observational human evaluation.
For each system, we asked participants to converse with it for at least 10 turns (5 from users and 5 from systems), and score every utterance from the system based on sensibleness and specificity. We totally evaluated 60 sessions for each system.

\begin{figure}[t]
  \centering
  \includegraphics[width=\linewidth]{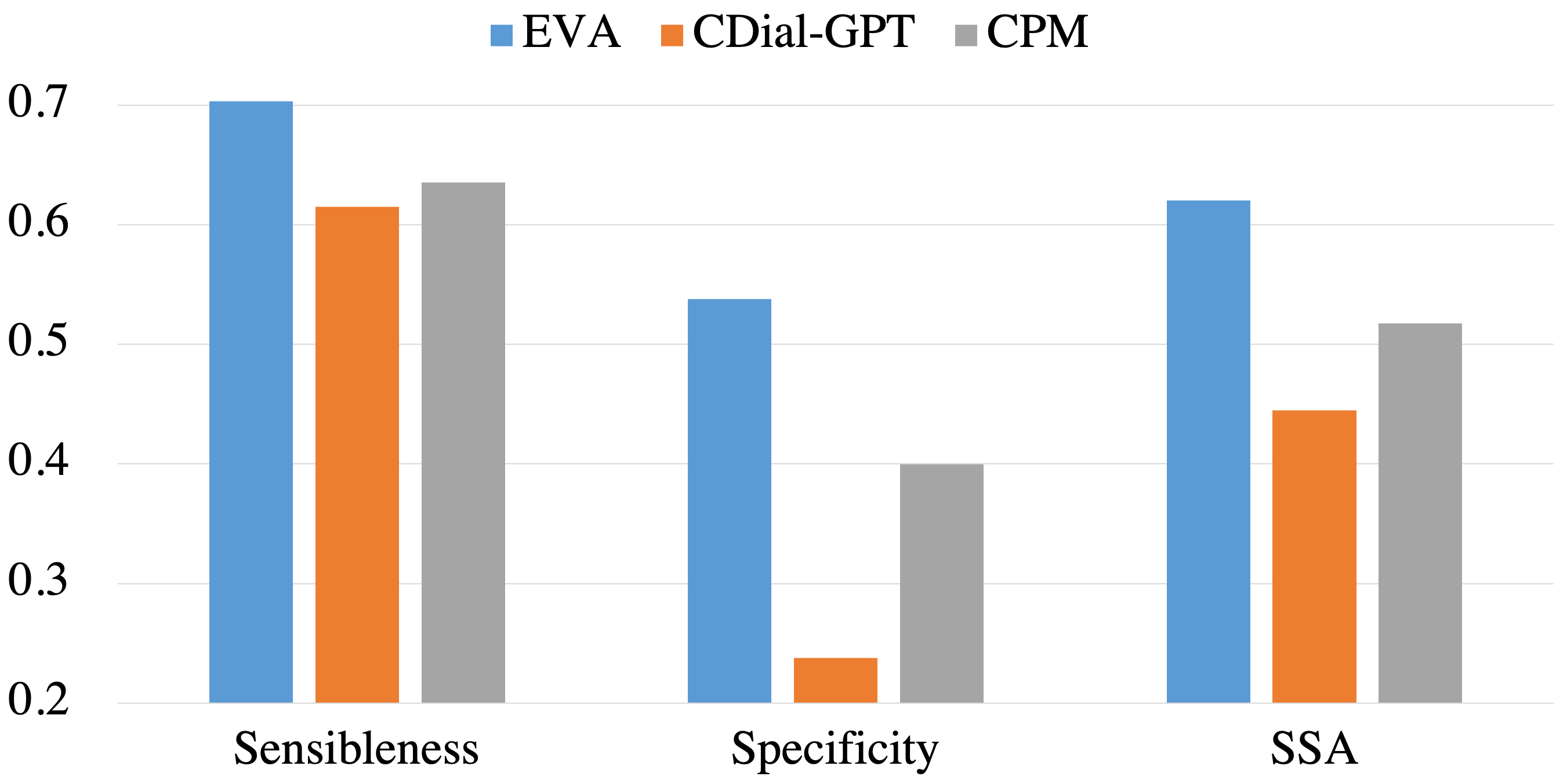}
  \caption{
  Results of interactive human evaluation.
  }
  \label{fig:inter}%
\end{figure}

The results are shown in Figure \ref{fig:inter}. We can observe that EVA obtains the highest sensibleness, specificity and SSA scores, showing the strongest ability of response generation in multi-turn human-bot interaction.

\begin{figure*}[]
  \centering
  \includegraphics[width=0.85\linewidth]{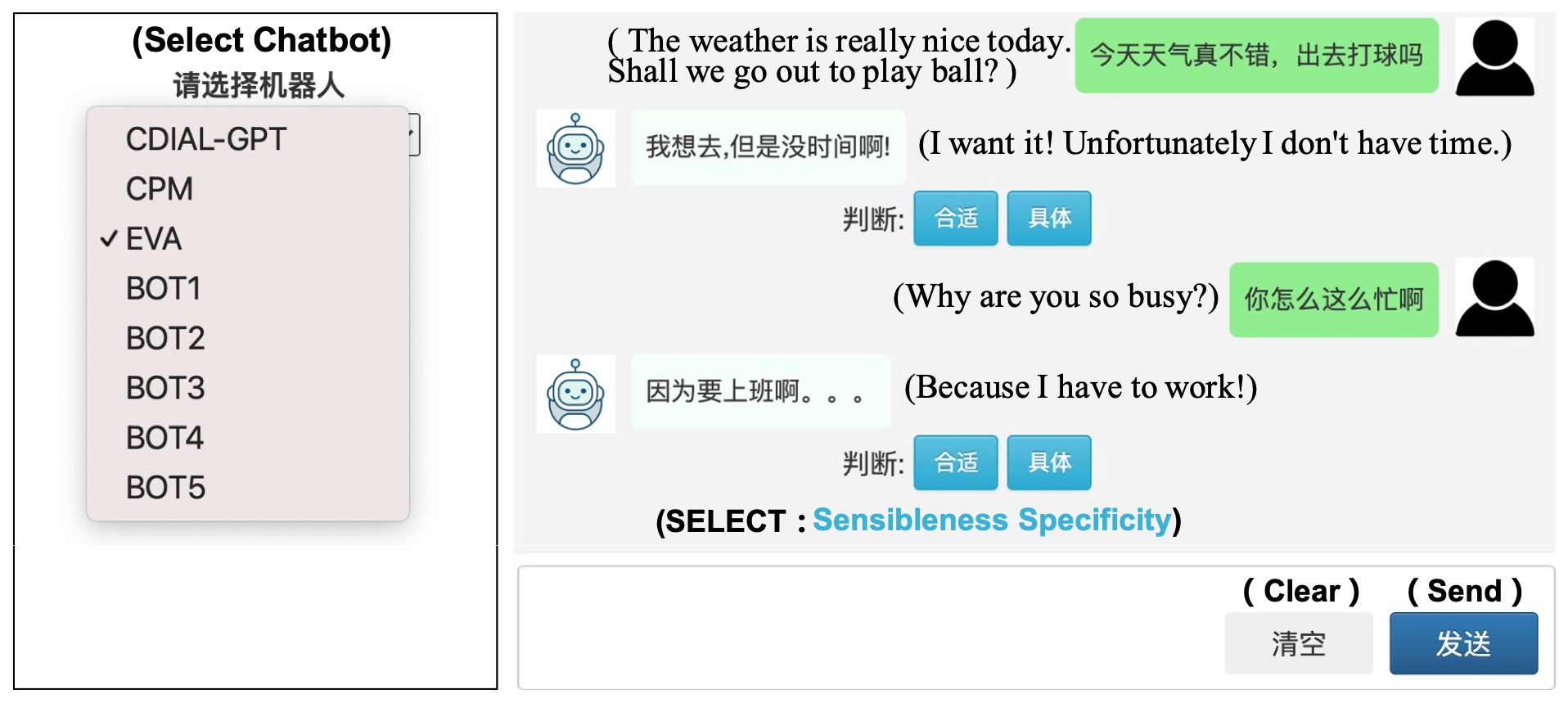}
  \caption{
  The interactive demonstration system of EVA.
  }
  \label{fig:interface}%
\end{figure*}%

\section{Case Study}
%%% model group

To intuitively show the generation ability of our model, we provide some generated cases in Table \ref{tab:staticcase} and Table \ref{tab:interactivecase}. We can observe from Table \ref{tab:staticcase} that EVA can generate more specific and relevant responses compared with the baselines. Table \ref{tab:interactivecase} shows a challenging case in interactive human evaluation, which demonstrates the strong ability of EVA in multi-turn human-bot interactions.

\begin{CJK*}{UTF8}{gbsn}

\begin{table} [!htp]
\centering
\footnotesize
\setlength{\tabcolsep}{1.0mm}{
\begin{tabu}{ll}
\toprule
Context: & 为什么感觉没睡醒 \\
& Why don't I feel awake?\\
\midrule
CDial-GPT: & 因为你没有睡醒 \\
& Because you are not awake. \\
CPM: & 我也不知道,反正就是感觉 \\
& I don't know. Anyway, it's just my feel. \\
%\hline
\midrule
EVA:  & 我还以为你不睡觉呢。这么晚了都 \\
& I thought you did not sleep. It's too late. \\
\bottomrule

\end{tabu}}
\caption{Examples of generated responses given the context in the test set.}
\label{tab:staticcase}
\end{table}

\end{CJK*}

\section{Interactive Demo}

%%% framework group
In addition to the above dialogue system, we also release an interactive demonstration system that enables researchers to converse with our system conveniently. It is an front-end interactive toolkit which can be easily built up through a JSON configuration file to communicate with the back-end dialogue system. In addition to the EVA model, developers can also deploy any other dialogue systems through the configuration file. As shown in the left side of Figure \ref{fig:interface}, there are 8 different dialog models.
By using the interactive demonstration system, users can communicate with the dialogue system and provide their ratings to the system response. Our main concern is the sensibleness and specificity of system response. As shown in the right side of Figure \ref{fig:interface}, the system response of the EVA model in the first turn is both sensible and specific, and the response of the second turn is sensible but not that specific. 

\begin{CJK*}{UTF8}{gbsn}

\begin{table} [!htp]
\centering
\footnotesize
\setlength{\tabcolsep}{0.8mm}{
\begin{tabu}{ll}
\toprule
Role & Response \\
\midrule
User: & 今天天气不错 \\
& The weather is nice today. \\
EVA: & 我这里下了一天雨,好凉快。你那儿呢? \\
& It rained all day here, so it's cool. How about you? \\
User: & 我这边多云，也挺凉快的 \\
& It's cloudy here, and it's also pretty cool. \\
EVA: & 是啊!不过现在又停啦~ \\
& Yeah! But it stops raining now. \\
User: & 雨停了可以出去玩儿呀 \\
& You can go out to play if it stops raining. \\
EVA: & 嗯呐,可惜没人一起去玩咯。 \\
& Well, it's a pity that no one is going to play together. \\
User: & 我们一起呗 \\
& Let's play together. \\
EVA: & 好哇!你啥时候回来?要不咱们约一下吧哈哈。 \\
& Wow! When do you come back? Or let's make an \\
& appointment, haha. \\
User: & 好的呀，我明天过去找你 \\
& Okay, I will come to see you tomorrow. \\
EVA: & 哈哈哈行啊那就这么定啦\textasciitilde 晚安哈\textasciitilde \\
& Hahaha, let’s do it\textasciitilde Good night\textasciitilde \\
%User: & 晚安 \\
%& Good night. \\
%EVA: & 嗯哪,早点休息哦亲耐滴\textasciitilde \\
%& Um, rest early, oh dear\textasciitilde \\
\bottomrule

\end{tabu}}
\caption{An interactive example of multi-turn dialogue generation.}
\label{tab:interactivecase}
\end{table}

\end{CJK*}

\section{Conclusion}

We propose an open-domain Chinese dialogue system called EVA, which contains the largest Chinese pre-trained dialogue model with 2.8B parameters. To train EVA, we collect the largest Chinese dialogue dataset called WDC-Dialogue containing 1.4B context-response pairs, which is filtered by strict and effective rules. We conduct extensive experiments on automatic and human evaluation to show the effectiveness of our model.

\bibliography{custom}
\bibliographystyle{acl_natbib}

\end{document}